\documentclass[a4paper,conference]{IEEEtran}
\IEEEoverridecommandlockouts
\usepackage{cite}
\usepackage{amsmath,amssymb,amsfonts}
\usepackage{algorithmic}
\usepackage{graphicx}
\usepackage{textcomp}
\usepackage{xcolor}
\def\BibTeX{{\rm B\kern-.05em{\sc i\kern-.025em b}\kern-.08em
    T\kern-.1667em\lower.7ex\hbox{E}\kern-.125emX}}
    
\usepackage{booktabs}

\usepackage[hidelinks]{hyperref}
\usepackage{multirow}
\usepackage{float}
\usepackage[abbreviations]{glossaries-extra}
\glsdisablehyper

\glssetcategoryattribute{acronym}{indexonlyfirst}{true}

\newabbreviation{atr}{ATR}{Automatic Target Recognition}
\newabbreviation{cw}{CW}{Continuous-Wave}
\newabbreviation{fmcw}{FMCW}{Frequency-Modulated Continuous-Wave}
\newabbreviation{svm}{SVM}{Support Vector Machine}
\newabbreviation{uav}{UAV}{Unmanned Aerial Vehicle}
\newabbreviation{fgsm}{FGSM}{Fast Gradient Sign Method}
\newabbreviation{pgd}{PGD}{Projected Gradient Descent}
\newabbreviation{stft}{STFT}{Short-time Fourier Transform}
\newabbreviation{cvd}{CVD}{Cadence-Velocity Diagram}

\usepackage{fancyhdr}
    
\begin{document}

\title{Robustness of Deep Neural Networks\\for Micro-Doppler Radar Classification
\thanks{This work was funded by EPSRC under Grant EP/R513349/1.}
}

\author{\IEEEauthorblockN{
Mikolaj~Czerkawski,
Carmine~Clemente,
Craig~Michie,
Ivan~Andonovic,
Christos~Tachtatzis
}
                                    % ...

\IEEEauthorblockA{
Department of Electronic and Electrical Engineering, University of Strathclyde, Glasgow, G1~1XW, UK}
}

\fancyhf{}
\renewcommand{\headrulewidth}{0pt}
\fancyfoot[c]{}
\fancypagestyle{FirstPage}{
\lfoot{978-83-956020-4-7 \copyright 2022 Warsaw University of Technology}
\rfoot{IRS 2022, Gdańsk, Poland}
}

\maketitle
\thispagestyle{FirstPage}

\begin{abstract}

    With the great capabilities of deep classifiers for radar data processing come the risks of learning dataset-specific features that do not generalize well. In this work, the robustness of two deep convolutional architectures, trained and tested on the same data, is evaluated. When standard training practice is followed, both classifiers exhibit sensitivity to subtle temporal shifts of the input representation, an augmentation that carries minimal semantic content. Furthermore, the models are extremely susceptible to adversarial examples. Both small temporal shifts and adversarial examples are a result of a model overfitting on features that do not generalize well. As a remedy, it is shown that training on adversarial examples and temporally augmented samples can reduce this effect and lead to models that generalise better. Finally, models operating on cadence-velocity diagram representation rather than Doppler-time are demonstrated to be naturally more immune to adversarial examples.
    
\end{abstract}


\section{Introduction}

    Most physical targets of interest, such as humans or vehicles, are composed of several reflecting segments, moving at different velocities that evolve over time depending on the performed motion, often periodic, like walking, running, or propelling. The micro-Doppler signatures of these targets often contain enough discriminative features to perform various classification tasks, such as human activity recognition~\cite{glasgowdataset, Seyfioglu2018, Le2018, Jia2020}, or classification of drone targets~\cite{Pallotta2020, Clemente2021, Park2021}.

    The accuracy of micro-Doppler radar classification is dependent on the capabilities of the classifier model. To that end, a lot of recent works~\cite{Seyfioglu2018, Le2018, Jia2020} have focused on utilizing deep neural network classifiers for these tasks.
    
    Despite the growing interest in deep neural classifiers for micro-Doppler data, a relatively limited attention has been paid to detailed analysis of the classifier behavior. More often than not, general performance statistics, such as accuracy or F1 score are reported as an indication of the achieved performance, while model robustness is not examined. Here, this gap is explored by investigating the effect on the model performance for two types of sample augmentation, one that can naturally arise when deployed in the field (small temporal shifts) and one that is optimized in an adversarial manner to take advantage of the network weights and result in incorrect output. The analysis includes prediction stability, susceptibility to adversarial examples, the degree to which these examples transfer across models, and finally, the influence of the input signal representation. The results provide grounds for selecting training practices of micro-Doppler signature classifiers that can increase the reliability of the models.
    
    Section~\ref{sec:prob_setting} describes the experimental setup, where six different training settings are described. Each training setting is explored in a study, where the properties of the resulting classifier are investigated, as reported in Section~\ref{sec:analysis}. The paper concludes with a summary in Section~\ref{sec:conclusion}. 

\section{Training Process}
\label{sec:prob_setting}

    An open dataset for human activity recognition~\cite{glasgowdataset} is used for the experiments presented in this work. The dataset\footnote{Available at \url{http://researchdata.gla.ac.uk/848/}} contains 1,752 \gls{fmcw} radar signatures of 6 different activities. Each sample is preprocessed by integrating the range bins of a range-time map and computing the Doppler-time representation as the \gls{stft} of the resulting temporal signal with a 64-point Blackman window and an overlap of 48. This is then interpolated to the network input size of 128 by 128, by using 128 spectral bins and downsampling in the temporal dimension. The dataset is split into training, validation, and test subsets, with ratios 50\%, 25\%, and 25\% in a stratified manner to preserve the proportions for each activity. The same split is maintained for all experiments in this work.
        
    \begin{figure*}
        \centering
        \begin{tabular}{cc}
            \includegraphics[width=0.45\textwidth]{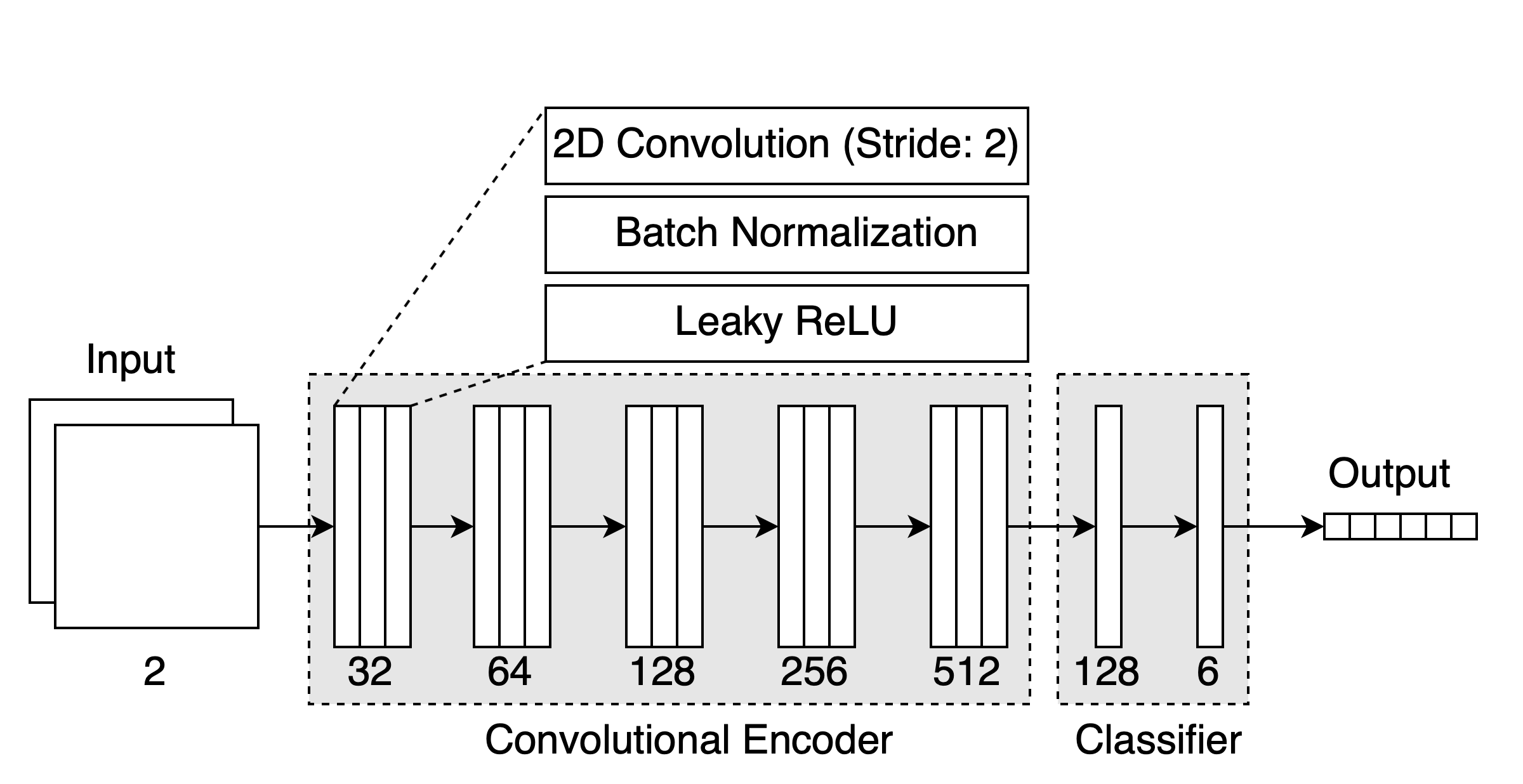} & \includegraphics[width=0.45\textwidth]{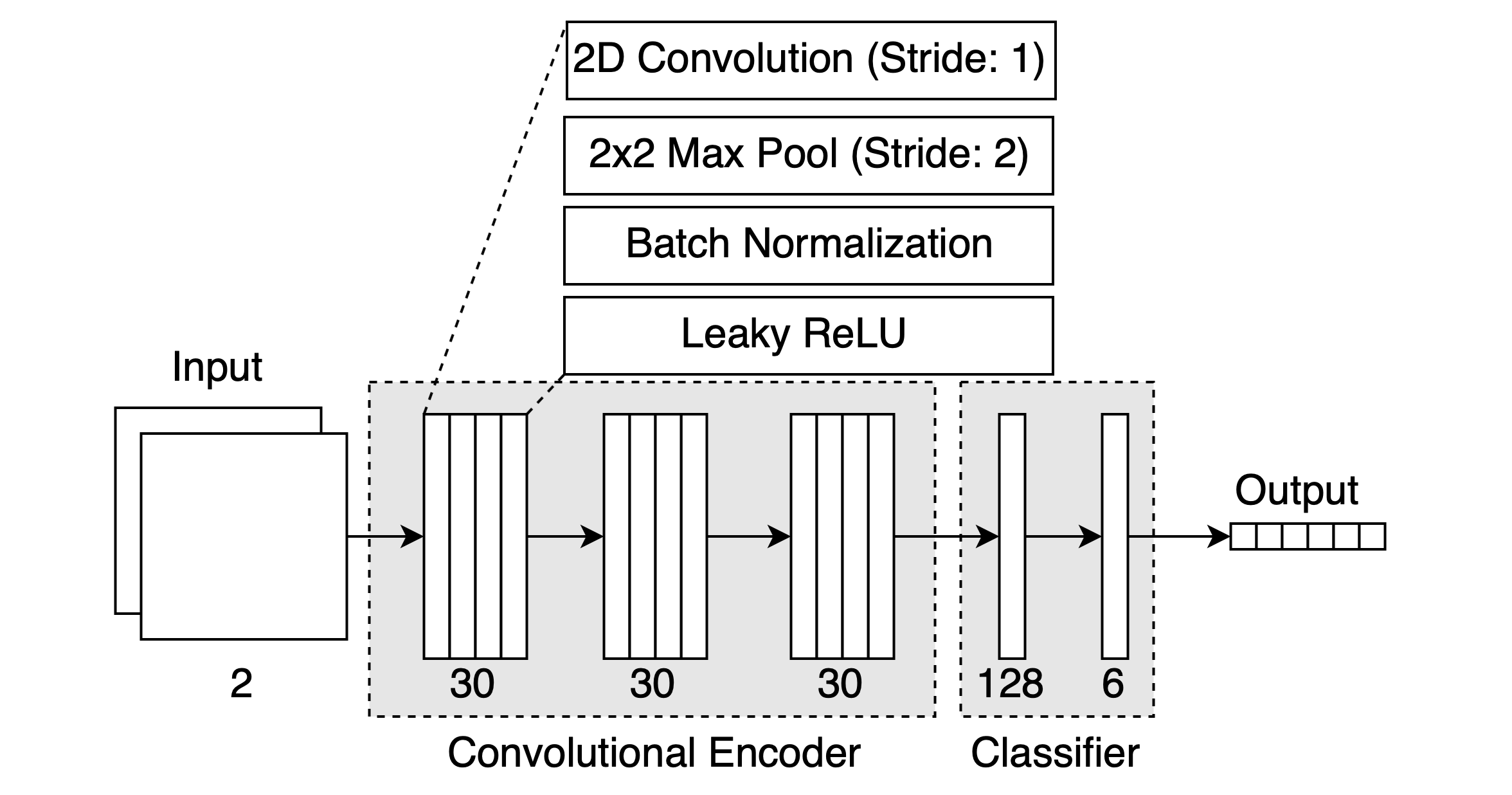}\\
            Model A & Model B
        \end{tabular}
        \caption{Two tested architectures. Both models use 9$\times$9 convolutional kernels.}
        \label{fig:arch_diagrams}
    \end{figure*}
    
    \begin{figure*}
        \centering
        \includegraphics[width=0.8\textwidth]{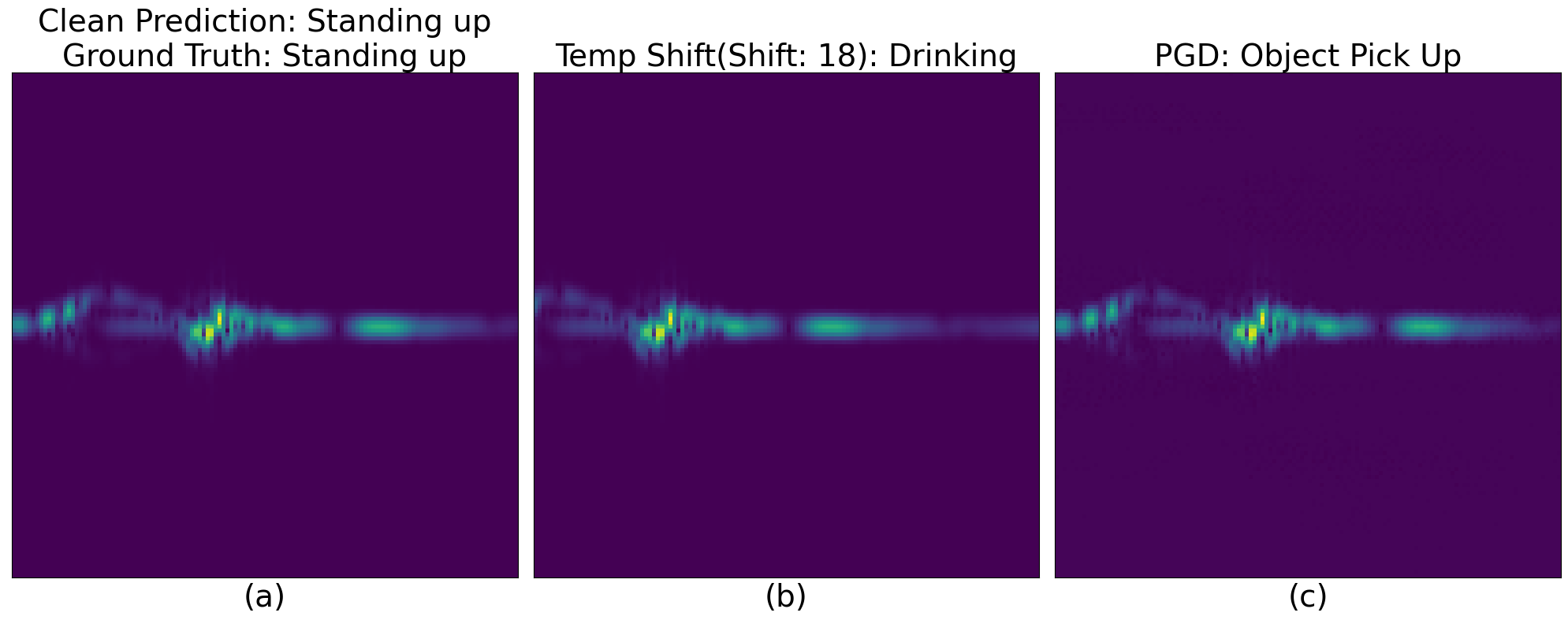}
        \caption{Without appropriate prevention mechanisms, models can express undesired sensitivity to changes with minimal semantic content. With the Model A trained in the standard manner, the clean sample of 'Standing Up` (a) is interpreted as 'Drinking` with a small temporal shift (b), and, as 'Object Pick Up` when a low-magnitude adversarial offset is added (c).}
        \label{fig:attack_examples}
    \end{figure*}
    
    As a case study, two different convolutional architectures are selected to test how adversarial examples transfer across independent models with varying architectures, both accepting input with two channels containing real and imaginary parts of the \gls{stft}. The first model, based on the architecture used in~\cite{Czerkawski2021,Czerkawski2022}, is shown in Figure~\ref{fig:arch_diagrams} as Model A. The second model, referred to as Model B, is based on one of the architectures used in~\cite{Seyfioglu2018} with exception to the activation function which is changed from ReLU to Leaky ReLU. For both models a classifier head with the same architecture is used, as shown in Figure~\ref{fig:arch_diagrams}.
    
    Here, two different methods of augmentation are explored, temporal shifts and adversarial examples. At the same time, the basic prevention mechanisms of adversarial training~\cite{Madry2018} and temporal augmentation are tested as defence strategies. Each of the two models is trained in four different training schemes: standard (S), adversarial (A), temporally augmented (T), and both adversarial and temporally augmented (A+T), resulting in a total of 8 trained models.
    
    The models are trained with cross-entropy loss using an Adam optimizer at the learning rate of 1e-3 for a maximum of 50 epochs; by this point, the validation would begin to increase for all conducted experiments. The model weights from the epoch yielding the lowest validation loss are used. Finally, Gaussian noise is added directly to the network input to yield SNR of 0 dB as a means of regularization.
    
    \begin{table}[!h]
        \centering
        \caption{Accuracies of the Tested models}
        \begin{tabular}{|ll|ccc|}
            \hline
            \multicolumn{2}{|c|}{Model}& \multicolumn{3}{c|}{Test Approach}\\
            \hline
            \multicolumn{1}{|c|}{Architecture} & Training & Standard  &  \multicolumn{1}{|c|}{PGD} & Temp Shift \\
            \hline
             & Standard & 0.86 & 0.00 & 0.76\\ 
            Model A & Adversarial (PGD) & 0.79 & 0.20 & 0.58\\ 
             & Temporal Aug & 0.91 & 0.04 & 0.84 \\ 
             & A+T & 0.90 & 0.32 & 0.81 \\ 
             \hline
             & Standard & 0.89 & 0.00 & 0.77 \\ 
            Model B & Adversarial (PGD) & 0.84 & 0.40 & 0.78 \\ 
             & Temporal Aug & 0.78 & 0.00 & 0.70 \\ 
             & A+T & 0.84 & 0.35 & 0.74 \\ 
             \hline
        \end{tabular}
        \label{tab:accuracy}
    \end{table}
    
    Table~\ref{tab:accuracy} shows that the models trained in the standard fashion suffer from significantly compromised performance for small temporal shifts (a few milliseconds), and are susceptible to adversarial examples with an accuracy dropping to 0.00. The three test approaches include a)~predicting on the original test samples (Standard) b)~predicting on adversarially augmented samples using \gls{pgd}~\cite{Madry2018} (20 steps, clipping values outside of -0.1 and 0.1), and c)~predicting on worst-case temporally shifted samples; in the range between 0 and 20 (Temp Shift). Examples of these augmentations are shown in Figure~\ref{fig:attack_examples}. The larger Model A appears to perform better with temporal augmentation, while Model B experiences reduced performance for both original and temporally shifted samples, which could be attributed to the limited capacity of the architecture. In terms of adversarial robustness, both models appear to benefit from the adversarial training practice.

\section{Analysis}
\label{sec:analysis}

    
    \begin{figure*}[!h]
        \centering
        \includegraphics[width=\textwidth]{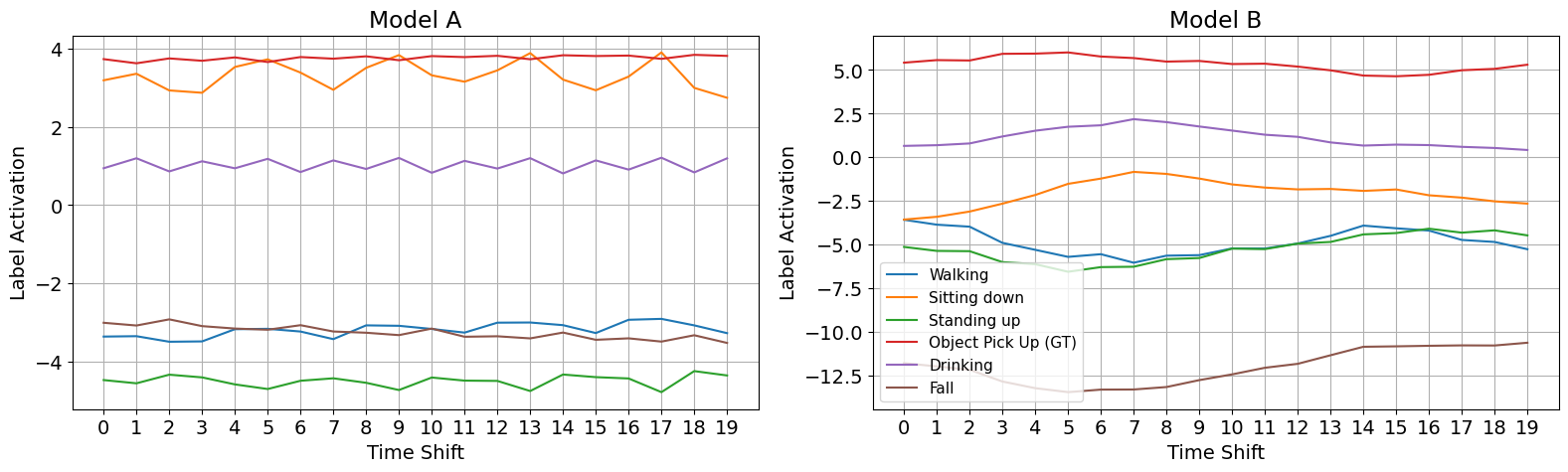} 
        \caption{Confidence Response to Temporal Shifts for an Object Pick Up Sample. Models trained in a standard manner.}
        \label{fig:temp_sample_1}
    \end{figure*}.
        
    \begin{figure*}[!t]
        \centering
        \includegraphics[width=\textwidth]{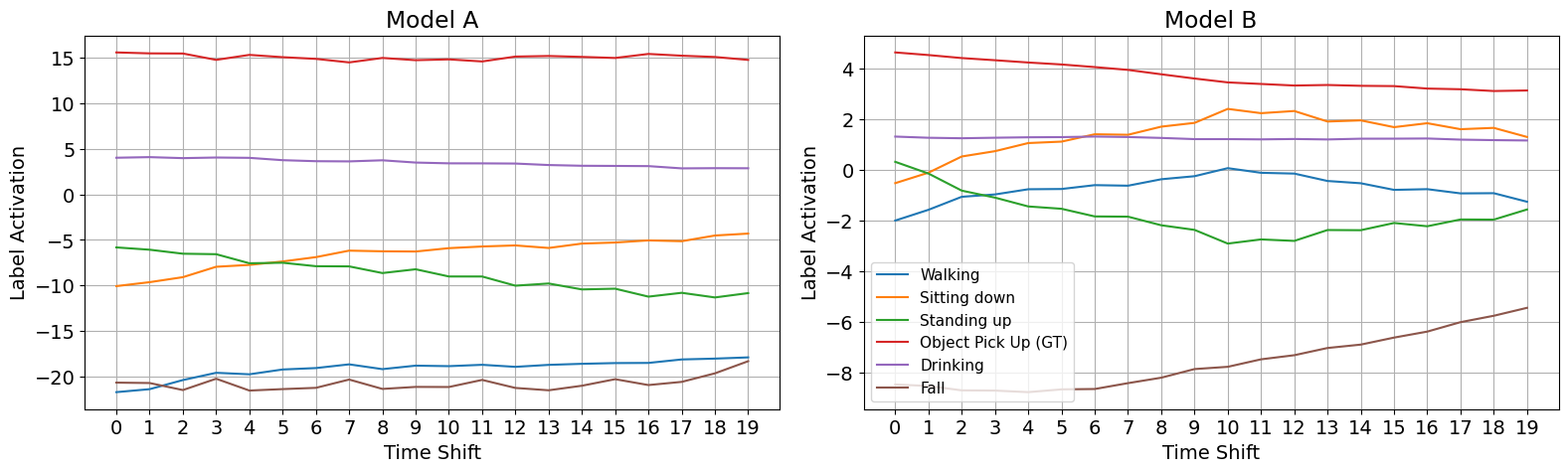}
        \caption{Confidence Response to Temporal Shifts for an Object Pick Up Sample. Models trained with temporal augmentation.}
        \label{fig:temp_sample_2}
    \end{figure*}

    The analysis investigates the generalization capability of the trained classifiers in several respects. First, the extent of translational invariance is studied across both temporal and Doppler dimensions of Doppler-time representations in subsection~\ref{subsec:translation}. 
    Further insight into susceptibility to adversarial examples is provided in subsection~\ref{subsec:adv_attacks}. These can be more harmful and, while less likely to occur, they can be used as an indicator of model robustness, along the lines of earlier works on the topic~\cite{Ilyas2019}.
    Finally, the third stage of the analysis investigates the effect the input representation on the classifier robustness by exploring models with \gls{cvd} input instead of Doppler-time.
    
    \subsection{Translational Invariance}
        \label{subsec:translation}
        Ideally, small shifts in the temporal dimension should have little effect on the predicted class label, assuming that the core information is distributed over the entire temporal scope. This assumption will be more or less valid depending on the sample and type of activity. To investigate the response of the networks to small shifts in time, test dataset samples have been obtained by interpolating the Doppler-time map to 148 spectra to allow a temporal shift up to 20 steps. To match the network input size, segments of 128 by 128 are extracted (where a single shift is about 4 ms or 8 ms for some longer samples).
        
        \begin{figure*}
            \centering
            \includegraphics[width=\textwidth]{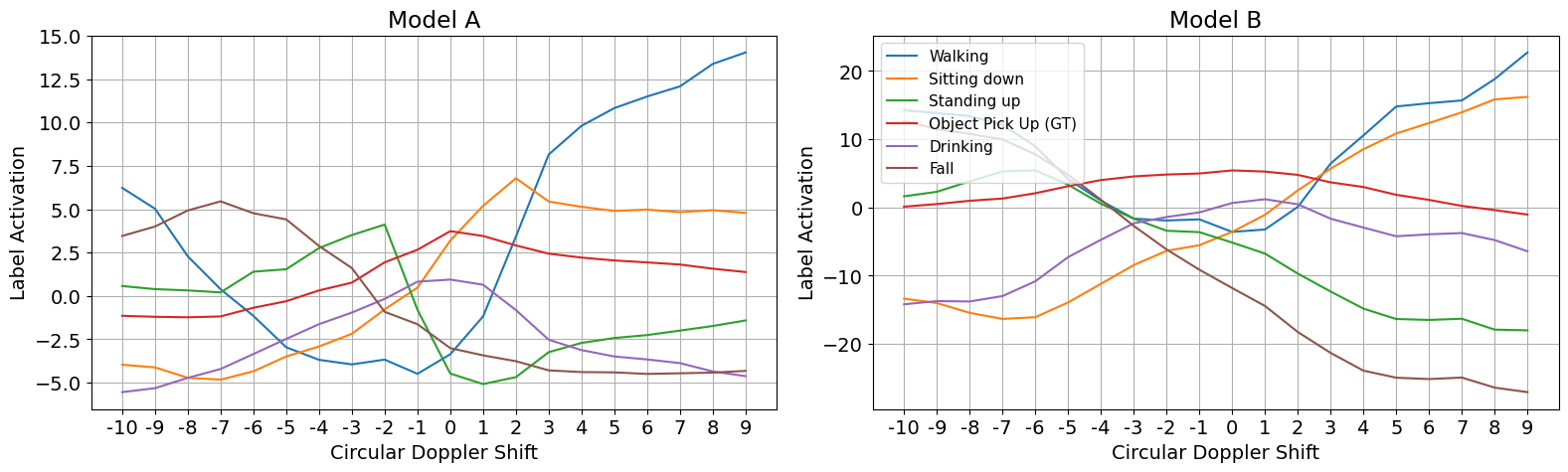}
            \caption{Confidence Response to Circular Doppler Shifts for an Object Pick Up Sample. Models trained in a standard manner.}
            \label{fig:dopp_sample}
        \end{figure*}
        
        Figure~\ref{fig:temp_sample_1} demonstrates how the output class confidence vector changes in response to small shifts in the temporal domain. The activation responses are considerably different for both models and their magnitudes are model-specific. Model B appears to be more invariant to temporal shifts, and the correct class is predicted for all tested shifts. On the other hand, the Model A exhibits stronger sensitivity to temporal shifts, and for some step values, such as 5, 9, 13, and 17, it outputs higher confidence for the 'Sitting Down' class than for the ground truth 'Object Pick Up'. A periodic relationship between the confidence and temporal shift in Model A has been observed for this sample, which could indicate overfitting of features in specific locations. Temporal sensitivity may be reduced by training with temporal augmentation, as shown in~Figure~\ref{fig:temp_sample_2} for Model A. However, with limited model capacity of Model B, it can also lead to lower accuracy, as discussed in Section~\ref{sec:prob_setting}.
        
        
        In Figure~\ref{fig:dopp_sample}, a similar effect is observed for shifts in the Doppler dimension. Unlike the temporal dimension, a shift of Doppler spectrum is readily interpretable. For instance, both models assign very high confidence to 'Walking' (blue line) as the sample is shifted further away from the null-velocity even though the signal pattern corresponds to an 'Object Pick Up' activity. This indicates that both networks are biased to detect 'Walking' based on Doppler offset rather than shape of the signature.  Interestingly, both models seem to be more confident about 'Walking' for a positive circular shift in the Doppler dimension. This could indicate a bias within the dataset of walking away from the sensor more than towards it. In the same way, the event of 'Fall' appears to be sensitive to motion towards the sensor significantly more. Other classes, appear to be highly correlated with the Doppler offset, with the 'Sitting Down' class consistently increasing in confidence for velocity offset suggesting motion away from the sensor, the 'Standing Up' class increasing for motion towards the sensor, and 'Drinking' confidence decreasing for velocities further away from the null. These observations highlight bias associated with the dataset collection and remit.
        
        \begin{figure*}
            \centering
            \includegraphics[width=0.99\textwidth]{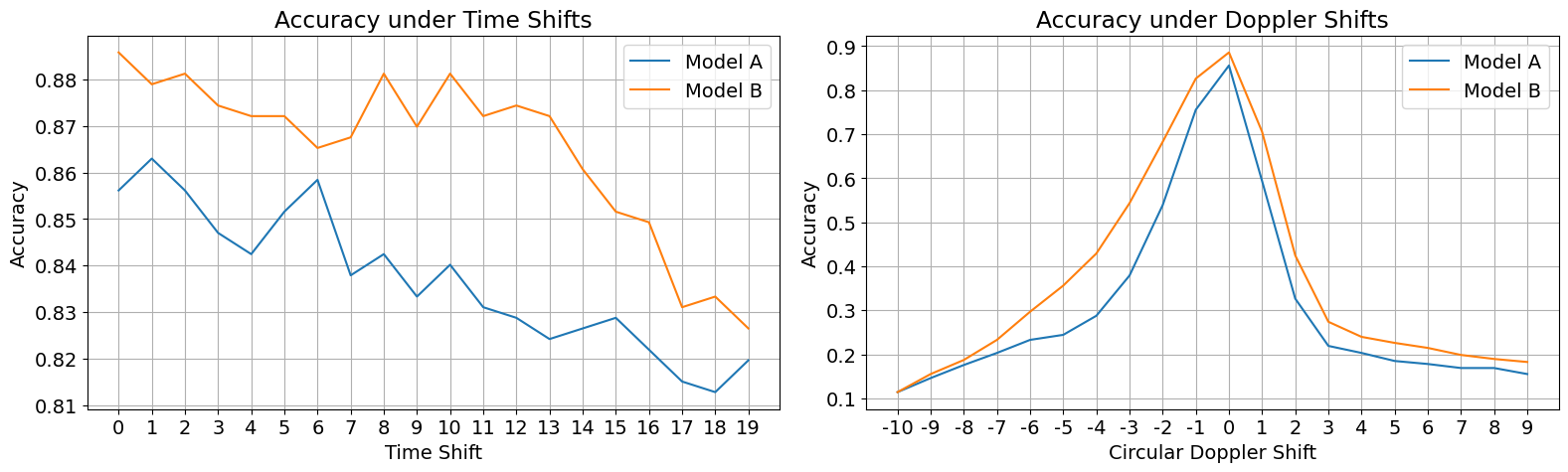}
            \caption{Accuracy response to a range of temporal and Doppler shifts of the input representations.}
            \label{fig:acc_drops}
        \end{figure*}
        
        \begin{table*}[]
            \centering
            \caption{Mean variance of the network response to shifts in temporal and Doppler dimensions for individual classes.}
            \begin{tabular}{|l | r r | r r | r r | r r|}
                \hline
                 &  \multicolumn{4}{c|}{Temporal Shift} & \multicolumn{4}{c|}{Doppler Shift}\\
                 \cline{2-9}
                Class & \multicolumn{2}{c|}{Model A} & \multicolumn{2}{c|}{Model B} & \multicolumn{2}{c|}{Model A} & \multicolumn{2}{c|}{Model B} \\
                \cline{2-9}
                 & \multicolumn{1}{c|}{Total} & GT & \multicolumn{1}{c|}{Total} & GT  & \multicolumn{1}{c|}{Total} & GT  & \multicolumn{1}{c|}{Total} & GT\\
                \hline
                Walking & 0.34 & 1.08 & 2.44 & 5.85 & 40.58 & 38.41 & 81.81 & 61.76\\
                Sitting down & 0.21 & 0.11 & 1.71 & 0.43 & 17.30 & 19.75 & 95.46 & 112.52\\
                Standing up & 0.25 & 0.54 & 1.31 & 0.59 & 7.29 & 12.89 & 86.05 & 81.51\\
                Object Pick Up & 0.08 & 0.04 & 0.47 & 0.63 & 2.47 & 2.76 & 3.54 & 4.75\\
                Drinking & 0.14 & 0.02 & 0.75 & 0.11 & 4.12 & 6.27 & 15.46 & 25.60\\
                Fall & 0.47 & 3.36 & 3.89 & 24.99 & 20.49 & 28.10 & 140.64 & 89.98\\
                \hline
            \end{tabular}
            \label{tab:conf_var}
        \end{table*}
        
        
        Deciding whether offsets in the Doppler dimension should be disregarded or used as predictive features can be difficult and depends on the definition of the activity. For some classes, the net velocity can be useful for classification for the majority of events, but narrow cases can arise, such as when 'Walking' is performed on a treadmill and with no net velocity. In these cases, a model too dependent on the Doppler offset can fail to recognize the correct class. Hence, designers must acknowledge the relationship between the class definition and development to avoid unexpected behaviour.
        
        To quantitatively measure the effect of the temporal and Doppler shifts of the input representations, the mean label activation variance for each class is computed for (a) all the samples in the test dataset (Total) (b) the subset of samples for which that class is the ground truth (GT) as shown in Table~\ref{tab:conf_var}. The mean level activation variance has been computed for a range of temporal and Doppler circular shifts in an inclusive range between 0 and 20, and -10 and 10, respectively. The larger the mean variance, the more dependent a given class on the positioning in the temporal or Doppler dimension is.
        
        Two classes exhibit high sensitivity to temporal offsets across models, and both of these classes, 'Walking' and 'Falling', are associated with non-stationary motion. A plausible hypothesis is that the most predictive features for these classes are specific fragments of the total signature occurring when a target begins to move, rather than the general shape of Micro-Doppler signature. This means that, for example, if the input contains a target walking without the transition from still, then the models trained on this non-augmented dataset can fail to recognize the activity.
        
        For Doppler shifts, the mean label activation variance is high for most classes, other than 'Object Pick-Up' and 'Drinking'. Coincidentally, all other classes contain large-scale motion and may be associated by the classifier with a specific velocity offset more than with a detailed pattern.
        
        Furthermore, the accuracy computed for each class for the tested range of shifts is plotted in Figure~\ref{fig:acc_drops}. In agreement with the earlier results, the accuracy drops significantly for Doppler shifts, but there is also a visible decay of performance for temporal shifts.            
    
    \subsection{Adversarial Examples}
        \label{subsec:adv_attacks}
    
        Deep neural networks are widely known to be susceptible to adversarial examples~\cite{Szegedy2014, Goodfellow2015} that take advantage of the high-dimensional input space to find a small input perturbation resulting in an incorrect prediction, with the change usually not apparent to the human eye (for example, constrained by a $\ell_1$ norm). In the last years, a popular employed interpretation, introduced in~\cite{Ilyas2019}, is that the adversarial gradients are effectively a property of data the network has been trained on, explaining why these examples can transfer well across models if they are trained on similar or the same data. 
        
        A common method for finding adversarial examples is known as \glsfirst{pgd} and it involves multistep optimization of the sample within its $\ell_{\infty}$ neighbourhood~\cite{Madry2018} and this method has been used to find adversarial examples for the entire test dataset. As shown in the PGD column of Table~\ref{tab:accuracy}, these examples still affect model performance even when adversarial training practice is employed, resulting in accuracy of 0.20 and 0.40 for Model A and B, respectively. For the standard training mode with no prevention measures, the accuracy under \gls{pgd} adversarial examples is 0.00.
        
        
        Following the interpretation of adversarial examples as features of the training data rather than the models alone, a natural question to ask is whether two models, with different architectures and trained independently in all aspects other than the training data, are susceptible to similar adversarial examples. Table~\ref{tab:pgd_transfer} shows the accuracy of Model A and B when evaluated under the standard dataset (No Attack) and on datasets containing adversarial examples optimized using the alternative model (Transfer Source) trained using the Standard (S), Adversarial Training (A), Temporal Augmentation (T) and combination of the last two (A+T). Both models suffer from compromised performance when fed with adversarial examples computed using the other architecture, regardless of training scheme employed.
        
        \begin{table}[h]
            \centering
            \caption{PGD Example Transferability with Doppler-Time Input}
            \begin{tabular}{|l l|c|c|c|c|c|}
                \hline
                \multicolumn{2}{|c|}{Model}& \multirow{2}{*}{No Attack}&\multicolumn{4}{c|}{Transfer Source}\\
                \cline{1-2}\cline{4-7}
                \multicolumn{1}{|c|}{Architecture} & Mode & & S & A & T & A+T\\
                \hline
                 & S & 0.86 & 0.39  & 0.50 & 0.24 & 0.45 \\ 
                Model A & A & 0.79 & 0.74 & 0.56 & 0.73 & 0.56 \\ 
                 & T & 0.91 & 0.23 & 0.55 & 0.15 & 0.52 \\ 
                 & A+T & 0.90 & 0.86 & 0.69 & 0.84 & 0.72\\ 
                \hline
                 & S & 0.89 & 0.41  &0.80 & 0.25 & 0.45\\ 
                Model B & A & 0.84 & 0.63 & 0.76 & 0.61 & 0.64 \\ 
                 & T & 0.78 & 0.25 & 0.80 & 0.27 & 0.46 \\ 
                 & A+T & 0.84 & 0.80 & 0.76 & 0.81 & 0.64\\
                \hline
            \end{tabular}
            \label{tab:pgd_transfer}
        \end{table}
        
    \subsection{Input Representation}
        \label{subsec:input_rep}
        
        A degree of temporal invariance can be achieved by transforming the Doppler-time representation to \gls{cvd}, the magnitude of Fourier transform in the temporal dimension. Without the phase, the representation is invariant to shifted in time of the complex oscillatory components within each Doppler bin. This approach has already been used in some past works on radar signature classification~\cite{Molchanov2012, Bjorklund2012, Miller2013, Pallotta2014, Clemente2015}, however, it appears to be less prominent in the recently published frameworks relying on deep neural networks~\cite{Seyfioglu2018, Le2018, Jia2020}.
        
        One of the capabilities of deep networks is learning the necessary transformations of data that minimize the output prediction error. Furthermore, a common approach is to keep as much information in the samples as possible and let the network extract whatever features allow it to minimize the error. Removal of phase information in \gls{cvd} is in odds with this philosophy. However, as it is shown in this work, these models with high capacity can easily learn non-robust features of the training dataset, which yield high accuracy on the test dataset but perform poorly when deployed in field. For this reason, adjusting the input representation in a way that enforces more prior knowledge about the task can be beneficial. In this case, \gls{cvd} representation can increase the temporal invariance and force the network to learn a more robust set of features.
        
        In Table~\ref{tab:accuracy_cvd}, it is demonstrated that significant robustness to adversarial examples and temporal shifts can be achieved by transforming the Doppler-time input to \gls{cvd} before feeding into the network. Furthermore, the temporal augmentation practice appears to have limited effect, suggesting that the spectra for \gls{cvd} do not change significantly for the tested range of temporal shifts. This, in turn, indicates that the crucial information for prediction is not concentrated only in the beginning of the recordings, and that Doppler-time classifiers could potentially be match the performance of the \gls{cvd}-based networks with additional training techniques employed.
    
        \begin{table}[h]
            \centering
            \caption{Accuracies of the Tested models with CVD Input}
            \begin{tabular}{|ll|ccc|}
                \hline
                \multicolumn{2}{|c|}{Model}& \multicolumn{3}{c|}{Test Approach}\\
                \hline
                \multicolumn{1}{|c|}{Architecture} & Training & Standard  &  \multicolumn{1}{|c|}{PGD} & Temp Shift \\
                \hline
                & Standard & 0.88 & 0.58 & 0.85 \\ 
                Model A & Adversarial (PGD) & 0.87 & 0.81 & 0.83\\ 
                 & Temporal Aug & 0.88 & 0.54 & 0.85\\ 
                 & A+T & 0.87 & 0.81 & 0.83 \\ 
                \hline
                 & Standard & 0.85 & 0.27 & 0.86\\ 
                Model B & Adversarial (PGD) & 0.87 & 0.80 & 0.86\\ 
                 & Temporal Aug & 0.87 & 0.62 & 0.83\\ 
                 & A+T & 0.88 & 0.81 & 0.86 \\ 
                \hline
            \end{tabular}
            \label{tab:accuracy_cvd}
        \end{table}
        
        Furthermore, the \gls{pgd} examples learned by the models do not transfer well, as shown in Table~\ref{tab:pgd_transfer_cvd} and the impact from transferred adversarial examples is limited. While decrease in accuracy is observed, it rarely goes below 0.80, meaning that the non-robust features learned by models operating on \gls{cvd} representation are less specific to the training data and could be attributed to other factors in the training routine.

        \begin{table}[h]
            \centering
            \caption{PGD Example Transferability with CVD Input}
            \begin{tabular}{|l c|c|c|c|c|c|}
                \hline
                \multicolumn{2}{|c|}{Model}& \multirow{2}{*}{No Attack}&\multicolumn{4}{c|}{Transfer Source}\\
                \cline{1-2}\cline{4-7}
                \multicolumn{1}{|c|}{Architecture} & Mode & & S & A & T & A+T\\
                \hline
                &S & 0.88 & 0.85 & 0.82 & 0.83 &  0.82 \\ 
                Model A & A & 0.87 & 0.85 & 0.84 & 0.84 & 0.83 \\ 
                &T & 0.88 & 0.84 & 0.81 & 0.84 & 0.80 \\ 
                &A+T & 0.87 & 0.85 & 0.83 & 0.84 & 0.83 \\ 
                \hline
                &S & 0.85 & 0.79 & 0.85 & 0.81 & 0.77\\ 
                Model B & A & 0.87 &0.78 & 0.84 & 0.81 & 0.84\\ 
                &T & 0.87 & 0.79 & 0.85 & 0.85 &  0.80\\ 
                &A+T & 0.88 & 0.84 & 0.83 & 0.86 & 0.83 \\ 
                \hline
            \end{tabular}
            \label{tab:pgd_transfer_cvd}
        \end{table}
    
\section{Conclusion}
\label{sec:conclusion}

    Deep neural networks trained on Doppler-time radar signatures learn a considerable amount of non-robust features. They express some sensitivity to small temporal offsets and significant sensitivity to Doppler spectrum shifts. This effect can be reduced by training a large enough model on temporally augmented samples, but considerable performance decline for samples with temporal offsets compared to original samples is still observed.
    
    Similarly, the models learn less apparent features of very low power that are specific to the training dataset and are unlikely to generalize to new samples, demonstrated by accuracy of 0.00 for adversarial examples (all samples misclassified), unless an adversarial training practice is employed. A good trade-off of accuracy and robustness is achieved when adversarial and temporally augmented samples are used in training.
    
    A more reliable method for robust classifiers proposed here involves training models on \gls{cvd} representation with discarded phase context. The tested models achieve similar performance with a degree of emergent robustness to both \gls{pgd} adversarial examples and temporal shifts.

\bibliographystyle{IEEEtran}
\bibliography{IEEEabrv,main.bbl}

\begin{thebibliography}{10}
\providecommand{\url}[1]{#1}
\csname url@samestyle\endcsname
\providecommand{\newblock}{\relax}
\providecommand{\bibinfo}[2]{#2}
\providecommand{\BIBentrySTDinterwordspacing}{\spaceskip=0pt\relax}
\providecommand{\BIBentryALTinterwordstretchfactor}{4}
\providecommand{\BIBentryALTinterwordspacing}{\spaceskip=\fontdimen2\font plus
\BIBentryALTinterwordstretchfactor\fontdimen3\font minus
  \fontdimen4\font\relax}
\providecommand{\BIBforeignlanguage}[2]{{%
\expandafter\ifx\csname l@#1\endcsname\relax
\typeout{** WARNING: IEEEtran.bst: No hyphenation pattern has been}%
\typeout{** loaded for the language `#1'. Using the pattern for}%
\typeout{** the default language instead.}%
\else
\language=\csname l@#1\endcsname
\fi
#2}}
\providecommand{\BIBdecl}{\relax}
\BIBdecl

\bibitem{glasgowdataset}
\BIBentryALTinterwordspacing
F.~Fioranelli, S.~A. Shah, H.~Li, A.~Shrestha, and J.~Le~Kernec,
  ``\BIBforeignlanguage{en}{Radar signatures of human activities},'' 2019.
  [Online]. Available: \url{http://researchdata.gla.ac.uk/id/eprint/848}
\BIBentrySTDinterwordspacing

\bibitem{Seyfioglu2018}
M.~S. Seyfioǧlu, A.~M. {\"{O}}zbayoǧlu, and S.~Z. G{\"{u}}rb{\"{u}}z, ``{Deep
  convolutional autoencoder for radar-based classification of similar aided and
  unaided human activities},'' \emph{IEEE Transactions on Aerospace and
  Electronic Systems}, vol.~54, no.~4, pp. 1709--1723, 2018.

\bibitem{Le2018}
H.~T. Le, S.~L. Phung, A.~Bouzerdoum, and F.~H.~C. Tivive, ``{Human Motion
  Classification with Micro-Doppler Radar and Bayesian-Optimized Convolutional
  Neural Networks},'' \emph{ICASSP, IEEE International Conference on Acoustics,
  Speech and Signal Processing - Proceedings}, vol. 2018-April, pp. 2961--2965,
  2018.

\bibitem{Jia2020}
M.~Jia, S.~Li, J.~L. Kernec, S.~Yang, F.~Fioranelli, and O.~Romain, ``{Human
  activity classification with radar signal processing and machine learning},''
  \emph{2020 International Conference on UK-China Emerging Technologies, UCET
  2020}, no. Cvd, 2020.

\bibitem{Pallotta2020}
L.~Pallotta, C.~Clemente, A.~Raddi, and G.~Giunta, ``{A Feature-Based Approach
  for Loaded/Unloaded Drones Classification Exploiting micro-Doppler
  Signatures},'' \emph{IEEE National Radar Conference - Proceedings}, vol.
  2020-September, 2020.

\bibitem{Clemente2021}
C.~Clemente, L.~Pallotta, C.~Ilioudis, F.~Fioranelli, G.~Giunta, and A.~Farina,
  ``{Chebychev moments based Drone Classification, Recognition and
  Fingerprinting},'' \emph{Proceedings International Radar Symposium}, vol.
  2021-June, pp. 1--6, 2021.

\bibitem{Park2021}
D.~Park, S.~Lee, S.~U. Park, and N.~Kwak, ``{Radar-spectrogram-based UAV
  classification using convolutional neural networks},'' \emph{Sensors
  (Switzerland)}, vol.~21, no.~1, pp. 1--18, 2021.

\bibitem{Czerkawski2021}
M.~Czerkawski, C.~Ilioudis, C.~Clemente, C.~Michie, I.~Andonovic, and
  C.~Tachtatzis, ``Interference motion removal for doppler radar vital sign
  detection using variational encoder-decoder neural network,'' in \emph{2021
  IEEE Radar Conference (RadarConf21)}, 2021, pp. 1--6.

\bibitem{Czerkawski2022}
------, ``A novel micro-doppler~coherence~loss for~deep~learning
  radar~applications,'' in \emph{European Radar Conference 2021 (EuRAD 2021)},
  2022, p. Forthcoming.

\bibitem{Madry2018}
A.~Madry, A.~Makelov, L.~Schmidt, D.~Tsipras, and A.~Vladu, ``{Towards deep
  learning models resistant to adversarial attacks},'' \emph{6th International
  Conference on Learning Representations, ICLR 2018 - Conference Track
  Proceedings}, pp. 1--28, 2018.

\bibitem{Ilyas2019}
A.~Ilyas, S.~Santurkar, D.~Tsipras, L.~Engstrom, B.~Tran, and A.~Madry,
  ``{Adversarial examples are not bugs, they are features},'' \emph{arXiv},
  2019.

\bibitem{Szegedy2014}
C.~Szegedy, W.~Zaremba, I.~Sutskever, J.~Bruna, D.~Erhan, I.~Goodfellow, and
  R.~Fergus, ``{Intriguing properties of neural networks},'' \emph{2nd
  International Conference on Learning Representations, ICLR 2014 - Conference
  Track Proceedings}, pp. 1--10, 2014.

\bibitem{Goodfellow2015}
I.~J. Goodfellow, J.~Shlens, and C.~Szegedy, ``{Explaining and harnessing
  adversarial examples},'' \emph{3rd International Conference on Learning
  Representations, ICLR 2015 - Conference Track Proceedings}, pp. 1--11, 2015.

\bibitem{Molchanov2012}
P.~Molchanov, J.~Astola, K.~Egiazarian, and A.~Totsky, ``{Classification of
  ground moving radar targets by using joint time-frequency analysis},''
  \emph{IEEE National Radar Conference - Proceedings}, no.~2, pp. 0366--0371,
  2012.

\bibitem{Bjorklund2012}
S.~Bj{\"{o}}rklund, T.~Johansson, and H.~Petersson, ``{Evaluation of a
  micro-Doppler classification method on mm-wave data},'' \emph{IEEE National
  Radar Conference - Proceedings}, pp. 0934--0939, 2012.

\bibitem{Miller2013}
A.~W. Miller, C.~Clemente, A.~Robinson, D.~Greig, A.~M. Kinghorn, and J.~J.
  Soraghan, ``{Micro-doppler based target classification using multi-feature
  integration},'' \emph{IET Conference Publications}, vol. 2013, no. 619 CP,
  2013.

\bibitem{Pallotta2014}
L.~Pallotta, C.~Clemente, A.~{De Maio}, J.~J. Soraghan, and A.~Farina,
  ``{Pseudo-Zernike moments based radar micro-Doppler classification},''
  \emph{IEEE National Radar Conference - Proceedings}, pp. 850--854, 2014.

\bibitem{Clemente2015}
C.~Clemente, L.~Pallotta, A.~{De Maio}, J.~J. Soraghan, and A.~Farina, ``{A
  novel algorithm for radar classification based on doppler characteristics
  exploiting orthogonal Pseudo-Zernike polynomials},'' \emph{IEEE Transactions
  on Aerospace and Electronic Systems}, vol.~51, no.~1, pp. 417--430, 2015.

\end{thebibliography}

\end{document}